\PassOptionsToPackage{utf8}{inputenc}
\documentclass{package/bioinfo}
\copyrightyear{2022} 
\pubyear{2022}

\access{}
\appnotes{}
\usepackage{amsmath}
\usepackage{multirow}
\usepackage[dvipsnames]{xcolor}
\usepackage{mathtools}
\usepackage{booktabs}
\usepackage{color, colortbl}
\usepackage{comment}

\begin{document}
\firstpage{1}
\subtitle{System Biology}

\title[CSI]{\textbf{CSI}: \underline{\textbf{C}}ontrastive Data \underline{\textbf{S}}tratification for \underline{\textbf{I}}nteraction Prediction and its Application to Compound-Protein Interaction Prediction}
\author[Kalia \textit{et~al}.]{Apurva Kalia\,$^{\text{\sfb 1}}$, Dilip Krishnan\,$^{\text{\sfb 2}}$ and Soha Hassoun\,$^{\text{\sfb 1,3,}*}$}
\address{$^{\text{\sf 1}}$Department of Computer Science, Tufts University, Medford, MA 02155, USA  \\
$^{\text{\sf 2}}$Google Research\\
$^{\text{\sf 3}}$Department of Chemical and Biological
Engineering, Tufts University, Medford, MA 02155, USA.}

\corresp{$^\ast$To whom correspondence should be addressed.}

\history{}

\editor{}

\abstract{\textbf{Motivation:}
Accurately predicting the likelihood of interaction between two objects (compound-protein sequence, user-item, author-paper, etc.) is a fundamental problem in Computer Science. Current deep-learning models rely on learning accurate representations of the interacting objects.  Importantly, relationships between the interacting objects, or features of the interaction, offer an opportunity to partition the data to create multi-views of the interacting objects.  The resulting congruent and non-congruent views can then be exploited via contrastive learning techniques to learn enhanced representations of the objects.\\
\textbf{Results:} 
We present a novel method, \textbf{C}ontrastive  \textbf{S}tratification for \textbf{I}nteraction Prediction (\textbf{CSI}), to stratify (partition) a dataset in a manner that can be exploited via Contrastive Multiview Coding (CMC) to learn embeddings that maximize the mutual information across congruent data views. CSI assigns a key and multiple views to each data point, where data partitions under a particular key form congruent views of the data. We showcase the effectiveness of CSI  by applying it to the compound-protein sequence interaction prediction problem, a pressing problem whose solution promises to expedite drug delivery (drug-protein interaction prediction), metabolic engineering and synthetic biology (compound-enzyme interaction prediction) applications.   
Comparing CSI with a baseline model that does not utilize  data stratification and contrastive learning, and show gains in Average Precision ranging from 13.7\% to 39\% using compounds and sequences as keys across multiple drug-target and enzymatic datasets, and gains ranging from 16.9\% to 63\% using reaction features as keys across enzymatic datasets.
\textbf{Availability:} Code and dataset available at https://github.com/HassounLab/CSI\\
\textbf{Contact:} \href{soha.hassoun@tufts.edu}{soha.hassoun@tufts.edu}\\
\textbf{Supplementary information:} Available at \textit{Bioinformatics} online}

\maketitle

\section{Introduction}\label{intro}
Predicting the likelihood of interaction between two objects (e.g., user-item, spectator-movie, author-paper, label-image, compound-protein, and other pairs) is a fundamental problem in Computer Science. Recommender systems, for example, utilize methods based on matrix-factorization to predict unknown interactions between users and items \citep{xue2017deep,he2017neural}. In network graphs, link prediction methods can anticipate potential connections between two collaborators, or authors and papers \citep{vamathevan2019applications}. Image captioning is achieved by recognizing objects within an image and characterizing interactions among them\citep{yao2018exploring}. Predicting the interaction between a compound and protein sequence elucidates drug-protein interactions \citep{bagherian2021machine} and promiscuous enzymatic activities on substrates \citep{visani2021enzyme}. Across various tasks, the success of interaction prediction hinges on learned representations of the interacting objects, as high-quality representations capture key features of interest. 
Multiple strategies have been developed in the machine learning literature to generate compressed representations of data \citep{bengio2013representation, hinton2011transforming, goodfellow2020generative}. Importantly, the availability of multi-modal data that represent different aspects of the same object creates opportunities for multi-view learning techniques \citep{li2018survey}, which have proven to be a powerful way to learn representations, especially in the computer vision literature \citep{radford2021learning,tian2020contrastive}.
Some such techniques attempt to minimize the distances between congruent (same-object) views, while others  contrast congruent and non-congruent views of the data to push away embeddings of differing data points. 

When addressing the interaction prediction problem, multi-modal representation learning can be applied on each object involved in the interaction. In this case, each object is embedded within its own latent space. In some tasks, deriving congruent data views is a common place task, e.g., image cropping, chrominance, and luminance for image-related tasks. However, in other cases, identifying congruent multi-views of data is challenging or non-trivial (e.g., drugs, disease, etc). To address this issue, and to further improve on representation learning for interactions, we use stratification (data partitioning) to generate multiple views of the data and to establish congruent and non-congruent  views. Contrastive learning methods can then be applied on the stratified data to enhance learning. 

More specifically, we explore  how the \textit{relationship between  two interacting objects} provides an  opportune data stratification strategy that allows representation learning in a joint latent space. Many-to-many interaction relationships among  objects allow  data to be stratified into congruent views for each object - the object itself is one view and all other objects related to it are another view.  For example, in a spectator-movie interaction scenario, a set of movies preferred by the spectator becomes an alternate view of the spectator.  Similarly, a set of spectators could provide an alternate view on the movie. Spectator and movie embeddings can then be learned in a joint latent space. Furthermore, ~\textit{features of the interaction itself} can provide alternative views on the interacting objects. For example, where and when the interaction occurs can provide information about movies and spectators. Rational stratification  of the training data enables generating congruent and non-congruent views of the objects and/or their interactions. We refer to this data stratification strategy as \textbf{C}ontrastive  \textbf{S}tratification for \textbf{I}nteraction Prediction, or CSI. 

To demonstrate the effectiveness of CSI, we  focus on the problem of compound-protein interaction prediction, a fundamental problem in biochemistry that is prominent in drug discovery (drug-protein interaction prediction) and in understanding and engineering metabolism (compound-enzyme interaction prediction). Related deep-learning methods broadly perform two  tasks: representation learning of compounds and of protein sequences, and using the learned representations to predict interactions. Molecular representations can be learned from molecular fingerprints \citep{feng2018padme, lin2020deepgs, lee2019deepconv} or learned on the corresponding molecular graphs using Graph Neural Networks (GNNs)  \citep{tsubaki2019compound, nguyen2021graphdta}.  Deep learning models such as CNNs \citep{lee2019deepconv}, and  transformers \citep{min2021pre, huang2021moltrans} are  used to generate embeddings on protein sequences. Interaction models however remain simple, where representations are  concatenated, with or without attention, to predict interaction likelihood. Unlike 3D docking simulations \citep{decherchi2020thermodynamics}, deep-learning models allow screening a large number of putative interactions efficiently.  In addition to its importance, the problem of compound-protein interaction prediction  was  selected to demonstrate the effectiveness of CSI because of rich available data on enzymatic interactions.  Compound-enzyme interactions are derived from known biochemical reactions, and therefore  information regarding the interaction itself allow us to evaluate CSI when stratifying based on interaction features. 

The core idea in CSI is intuitive. Each data point is assigned a "key" and  multiple views. When learning molecular representations, each "key" is the molecule itself, and the corresponding views are the molecule and a set (or subsets) of interacting sequences. Similarly, when learning sequence representations, the "key" is the sequence itself, and the corresponding views are the sequence and the set (or subsets) of interacting molecules. When stratifying by interaction feature, the "key" is the interaction feature (e.g., all reactions that perform a specific biotransformation such as the addition of carboxyl group), and three views of each reaction (or reaction group, if the key places multiple reactions within a strata) are readily available: reactant-product pairs associated with the reaction (View 1), compound-sequence pairs  (View 2), and  sequences that catalyze the reaction (View 3), where the compounds are either reaction substrates or products.  Other interaction features can also be selected as keys (e.g., reactions sharing homologous sequences). Views under the same key  form congruent views of the data, while views across different keys become  non-congruent views. Once  congruent and non-congruent data views are established, it is possible to apply any contrastive learning technique to learn the joint representation. In our case, we use Contrastive Multiview Coding (CMC) \citep{tian2020contrastive}, which 
simultaneously maximizes the mutual information present among the congruent views of the data while discarding features that are not shared among the views. Importantly, our work demonstrates the importance of view selection when applying contrastive learning \citep{tian2020makes}. 

We train and evaluate CSI models for three datasets. The BindingDB dataset \citep{gilson2016bindingdb}  contains purchasable drugs and their protein targets that exhibit an affinity higher than 10 $\mu$M, and is larger and more diverse than earlier drug-protein interaction datasets.
The  BRENDA  dataset is derived from the BRENDA database \citep{chang2021brenda}, which provides continued manual and automated curation on enzymes and compounds interacting with enzymes. The KEGG dataset is derived from the KEGG database \citep{kanehisa2021kegg}, which catalogues biochemical  reactions for a large set of  organisms.  
The contributions of this paper are:
\begin{itemize}
    \item A generalizable data stratification method, CSI, for view selection on interacting objects, where  stratification is applied either on each of the items involved in the interaction in the context of the other object, or on features of the interaction itself. 
    
    \item Congruent and non-congruent data views allow CSI to be paired with contrastive learning schemes, such as CMC, resulting in  learned embeddings suited for downstream tasks.
 
    \item Demonstrating how CSI applies to the compound-protein interaction prediction task for protein-drug and to enzyme-compound datasets. The latter dataset is rich in auxiliary interaction information that lends itself to  stratification on  interaction features.
 
     \item Showing that CSI significantly outperforms a baseline model that does not use CSI, 
     where average precision (AP) is improved by 
      18.2\% on the BindingDB dataset,
     39\% on the BRENDA dataset, 
     and 
     13.7\% on the KEGG dataset, when stratifying by compound and by sequence.
  When stratifying by reaction features for the KEGG dataset, an AP improvement of 16.9\% is achieved over baseline, thus outperforming stratification by compound and by sequence.

\end{itemize}

\section{Methods}
\subsection{Stratification on interaction data - congruent views of compounds and of protein sequences}\label{strat_inter}
\begin{figure}[tb]
    \centering
    \includegraphics[width=\linewidth,trim={0 0.1cm   0 0},clip]{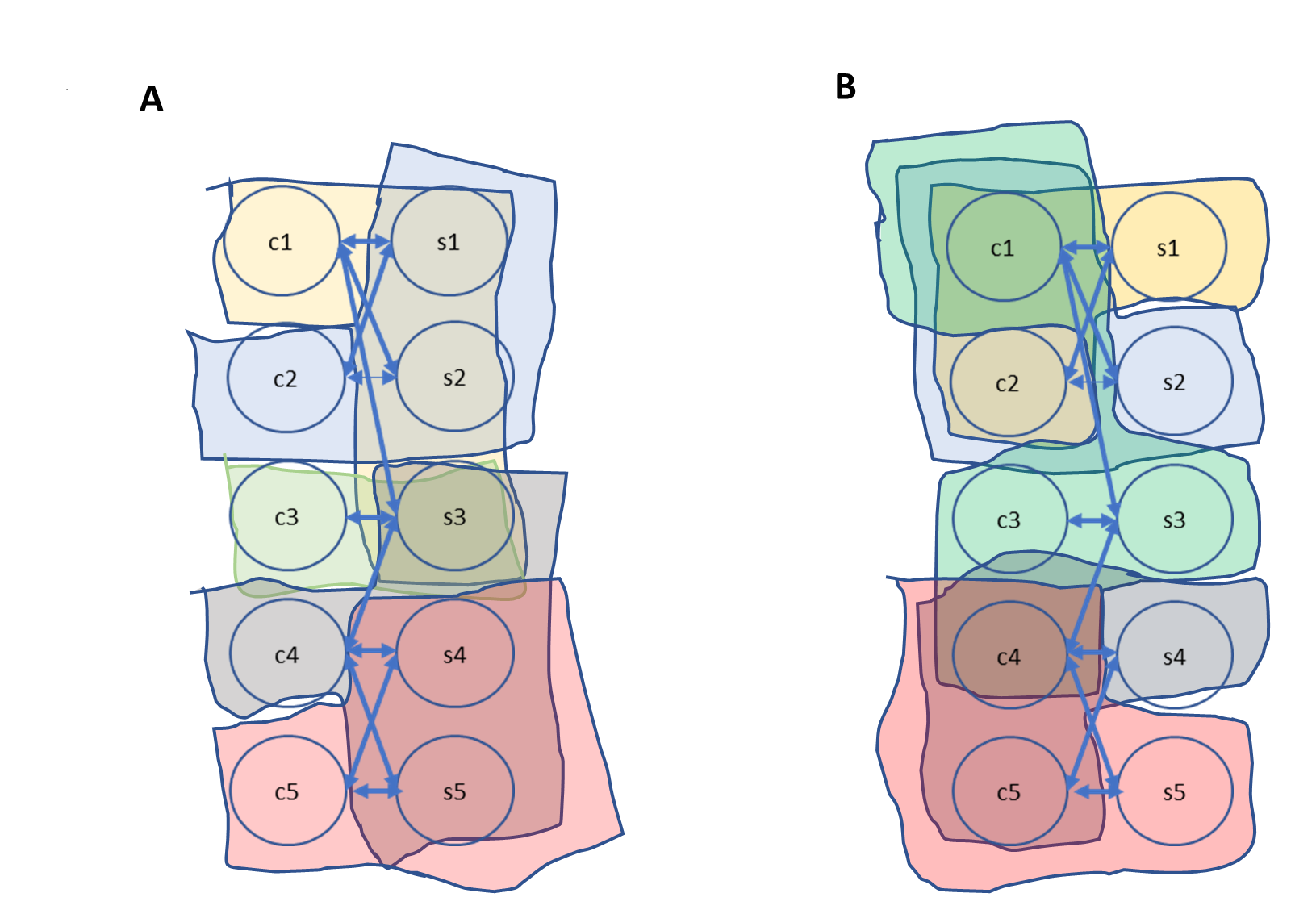}
    \caption{Many-to-many interactions between compounds and protein sequences allow data stratification by: (A) compound, and by (B) sequence.}
    \label{CSIfig:data_strat}
\end{figure}

An interaction dataset consists of compound-protein pairs known to interact. A compound may interact with multiple proteins, and  a protein may have interactions  with multiple compounds (Figure \ref{CSIfig:data_strat}). For data stratified using compounds as the key, the set of all protein sequences that interact with the given compound presents a view congruent with the compound.  Assuming a lock-and-key based binding model \citep{tripathi2017molecular}, the rationale for these views being congruent is that the interacting proteins have common features that enable  binding with the same compound. Subsets, or even pairs, of the protein sequences therefore offer a view that is congruent with the compound. To simplify our formulation and implementation, we use a pair of sequences as a congruent view of a compound. 
Assuming that  $I$ is the set of known interactions on compounds $C$ and a set of sequences $S$, the set of congruent views, $V_C$, for all compounds in $C$ is:
\begin{equation}
    V_C = \{[c, (s_i, s_j)],~s_i, s_j \in S, c \in C, \forall (c, s_i), (c, s_j) \in I\}
\end{equation}
where the square brackets denote views.

Stratification using sequences as keys is used to define congruent views for each sequence. A set of compounds, or a subset thereof, presents a congruent view of a sequence. Using a pair of compounds as a congruent view of a sequence, the set of congruent views, $V_S$, for all sequences in $S$ is: 
\begin{equation}
    V_S = \{[s, (c_i, c_j)], s \in S, c_i, c_j \in C, ~\forall (c_i, s), (c_j, s) \in I\}
\end{equation}

\subsection{Stratification on reaction data - congruent views on interaction features}\label{strat_react}
Compound-protein interactions within enzymatic datasets are associated with biochemical reactions. The auxiliary data available on the reactions can be used as keys for stratifying by interaction features (and not by compounds and sequences as presented in the prior section). Each reaction represents a set of reactants that undergo a biochemical transformation into a a set of products. Homologous enzyme sequences (e.g., enzymes from different organisms catalyze the same reaction) and multiple enzymes performing similar function  can catalyze the same reaction. A biochemical reaction, $b$, is  assumed to be bidirectional and can be represented as: 
\begin{equation}
    R \xleftrightarrow{\text{E}} P
\end{equation}
where $R$ is the set of reactants, $P$ is the set of products, and $E$ the set of enzyme sequences that catalyze the reaction.  A reaction can therefore be defined as, $b = \{R, E, P\}$, where the subscripts on $R, E,$ and $P$ are omitted for clarity. 
Each reaction therefore lends itself to three congruent views: a list of corresponding reactant-product pairs, a list of compound-sequence interactions, where a compound maybe a reactant or a product, and a list of catalyzing sequences. 
The set of  congruent views, $V_{B}$, for the set of biochemical reactions, $B$, is given by:
\begin{equation}
\begin{split}
    V_{B} = \{[(r_i, p_j), (c_i, s_k), (s_k, s_l)],\\
    \forall r_i \in R, p_j \in P, c_i \in R \cup P, s_k, s_l \in E,\\
    b=\{R,E,P\}, \forall  b \in B\}    
\end{split}\label{CSIeq:reaction}
\end{equation}

\subsection{CSI on  interacting objects}\label{CSI_arch}
CMC \citep{tian2020contrastive} arrives at data representations by learning embeddings for each view, and a function, $h_\theta$, that  discriminates a congruent pair among a set of non-congruent views  based on the learned embeddings. Once the embeddings are learned via encoders, their parameters are frozen and can be used for the downstream task. We adopt a similar methodology for CSI.  

\begin{figure*}[htbp]
    \centering
    \includegraphics[width=\textwidth]{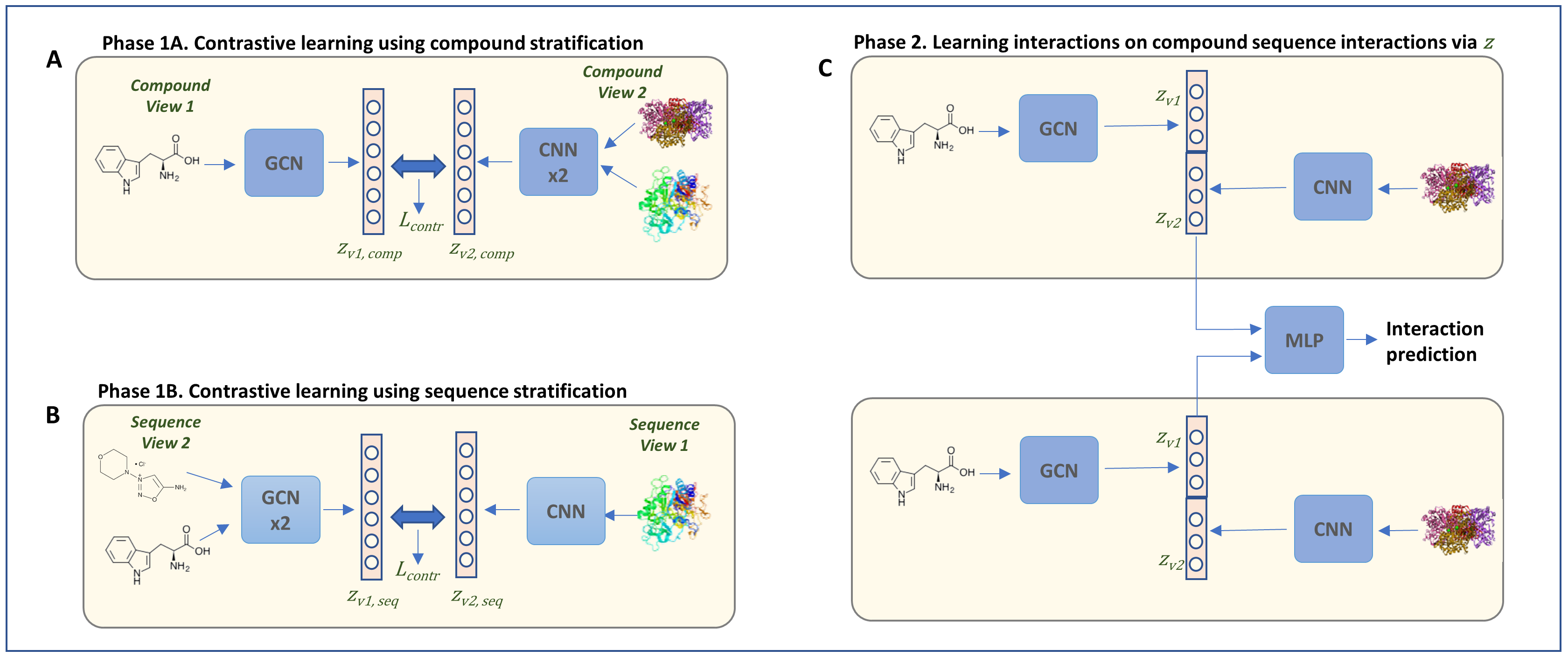}
    \caption{CSI model when stratifying each interacting object. (A) Phase 1A for compounds as keys – Compound representation, $z_{v1}$ is generated through a GCN and  sequence-sequence representation, $z_{v2}$ is generated using a Siamese CNN. (B) Similarly, in Phase 1B for sequences as keys, compound-compound representation,   $z_{v1}$, is generated through a Siamese GCN, while sequence representation, $z_{v2}$, is generated through a CNN. (C) 
    In Phase 2, the trained encoders from Phases 1A and 1B are fixed. The representations are concatenated to train an MLP for final prediction. }
    \label{fig:CSI2views}
\end{figure*}

CSI is trained in two phases (Figure \ref{fig:CSI2views}).  In the first phase, Phase 1A and 1B, we learn embeddings on compound views and, independently, on sequence views. In Phase 1A, for compound as key, each of the congruent views of the compounds, $V_c$, consist of one compound and two sequences. We therefore train encoders to generate embeddings for these two views, ensuring that they produce same-dimension embeddings. As compounds can be represented as graphs,  we utilize a Graph Neural Network (GCN) to encode the compounds:
\begin{equation}
    z_{v1, comp} = GCN(c)
\end{equation}
For the protein sequence, we use a 1-dimensional Convolutional Neural Networks (CNN) on the encoded FASTA \citep{lipman1985rapid} sequence, $F$, normalized to a fixed length (e.g., 1000). As we need to learn the representation of two sequences at-a-time to represent the compound, we utilize a Siamese CNN network that uses the same weights. The twin CNNs are trained in tandem on two encoded input sequences and compute the final embedding for the view, $z_{v2}$. That is, 
\begin{equation}   
    z_{v2, comp} = CNN(F_i) \oplus CNN(F_j)
\end{equation}
where $\oplus$ is the concatenation operation. 

In Phase 1B, for sequence as key, congruent views of a sequence, $V_s$, comprise one sequence, $s$, and two compounds, $c_i$ and $c_j$. To learn the embeddings for these views, we utilize a CNN for the encoded sequence, and a Siamese GCN network for the two compounds. That is,
\begin{equation}
    z_{v1, seq} = GCN(c_i) \oplus GCN(c_j)
\end{equation}
\begin{equation}
    z_{v2, seq} = CNN(s)
\end{equation}
Independent GCNs and CNNs are trained in Phase 1A and Phase 1B. 

The discriminator function, $h_\theta$, between embeddings for pairs of n-th and m-th objects from two views $v1$ and $v2$ is defined as in prior work  \citep{tian2020contrastive} as the cosine similarity between their embeddings modulated by a temperature parameter $\tau$:
\begin{equation}
    h_\theta(z^n_{v1}, z^m_{v2}) = exp(\frac{z^n_{v1}.z^m_{v2}}{\|z^n_{v1}\|.\|z^m_{v2}\|}.\frac{1}{\tau})\label{CSIeq:critic}
\end{equation}
where $\tau$ is a hyper-parameter that controls the importance of  non-congruent views in pushing the embeddings apart in the latent space. We define the contrastive loss over a batch of size $k$ as:
\begin{equation}
    L^{V_1, V_2}_{contrastive} = \frac{1}{k} \sum\limits^k_{n=1}\left[-\mathbb{E}[log\frac{h_\theta(z^n_{v1}, z^n_{v2})}{\sum\limits^k_{m=1, m \neq n}h_\theta(z^n_{v1}, z^m_{v2})}]\right]\label{CSIeq:loss}
\end{equation}
Defining the contrastive loss in the context of a batch facilitated the CSI implementation and avoided complex strategies to select the non-congruent views \citep{tian2020contrastive}. In essence, we select non-congruent views within a batch, instead of considering all possible non-congruent views within the entire dataset. As the contrastive loss $L^{V_1, V_2}_{contrastive}$ treats $V_1$ as the anchor view and iterates over $V_2$, it is not symmetrical.  We can similarly anchor $V_2$ and iterate over $V_1$ to arrive at $L^{V_2, V_1}_{contrastive}$.  The total contrastive loss  \citep{tian2020contrastive}, giving equal weight to both views, is then,
\begin{equation}
    L(V_1, V_2) = L^{V_1, V_2}_{contrastive} + L^{V_2, V_1}_{contrastive}
\end{equation}

Once the encoders are trained to minimize the loss, their parameters are frozen during Phase 2. The interaction predictor is an MLP neural network that utilizes the learned embeddings for the  compound views, and for the sequence views. The interaction predictor is trained on known positive interactions and on negative interactions, which consists of  randomly selected compound-sequence pairs.  For the contrastive loss (Equation \ref{CSIeq:loss}), views from different keys within a batch (e.g., one compound and two sequences for the compound-based stratification) are taken as non-congruent, while for the final predictor training, randomly selected compound-sequence pairs are taken as  negative data. The final predictor $\hat{y}$ is given by,
\begin{equation}
    \hat{y} = MLP((z_{v1,comp} \oplus z_{v2,comp}) \oplus (z_{v1,seq} \oplus z_{v2,seq}))\label{CSIeq:final}
\end{equation}
The prediction loss is the cross entropy loss between $\hat{y}$ and the ground truth $y$ weighted by the ratio of negative to positive labeled data.

\subsection{CSI on interaction features}\label{CSI_arch2}
\begin{figure*}[htbp]
    \centering
    \includegraphics[width=\textwidth]{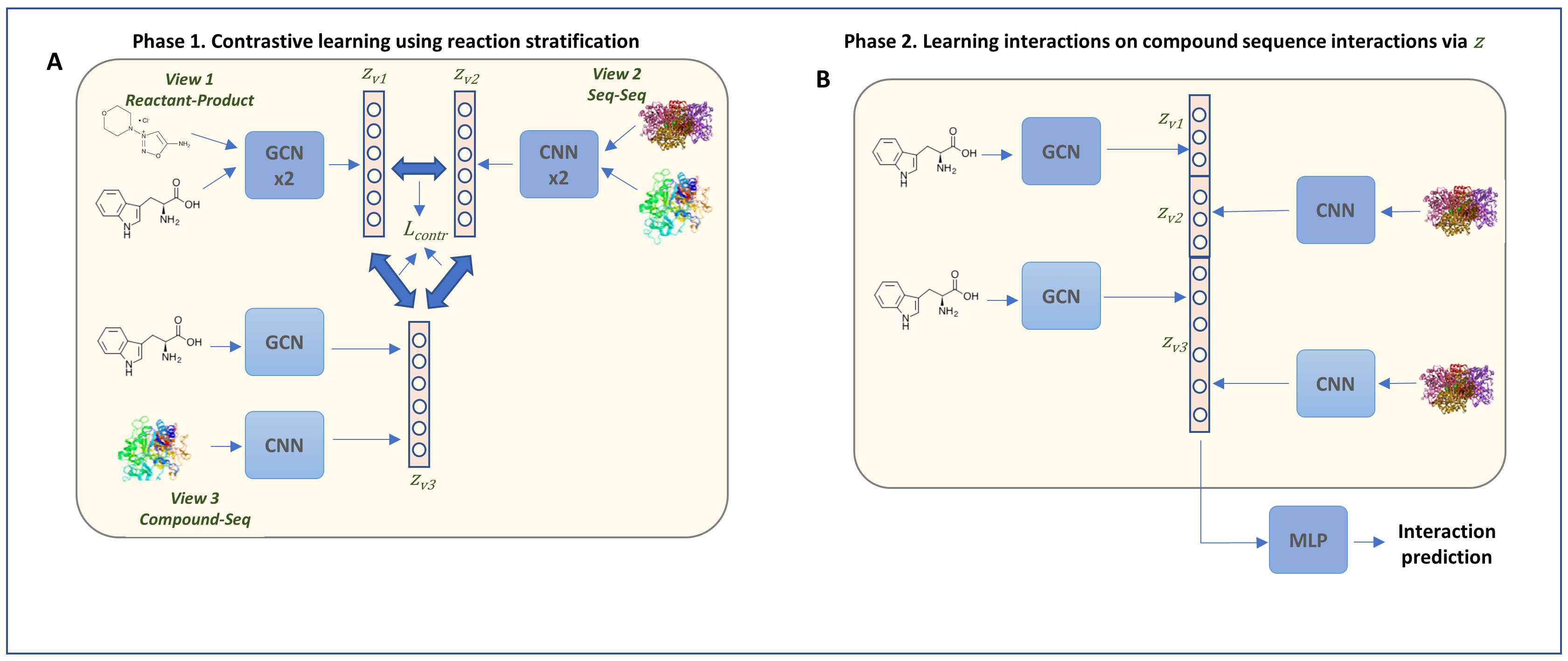}
    \caption{CSI model when stratifying by interaction feature. (A) Phase 1. Contrastive loss is applied to the three data views: compound-compound pairs, compound-sequence pairs, and sequence-sequence pairs to generate three embeddings, $z_{v_1}$,  $z_{v_2}$, and  $z_{v_3}$.
 (B) Phase 2. Trained encoders from Phase 1 are used to generate representations for compounds and sequences. These representations are concatenated together to train an MLP for the final prediction.}
    \label{fig:CSI3views}
\end{figure*}

When data is keyed by interaction features (Figure \ref{fig:CSI3views}), we  apply contrastive loss on three data views: a set of compound-compound pairs representing substrates-products, a set of paired compound-sequences and a set of sequences. The framework of CSI can be easily adapted to maximize the mutual information across the three views, as was suggested for CMC \citep{tian2020contrastive}, and to perform interaction prediction on the concatenated learned embeddings. In the first phase of CSI, Siamese GCN and CNN networks are used to learn the compound-compound and sequence-sequence embeddings, and a GCN-CNN are used to learn the embeddings for the compound-sequence embeddings.  The contrastive loss is  calculated pairwise, over all the views, as defined previously (Equation \ref{CSIeq:loss}). In the second phase,  encoder parameters are fixed, and the embeddings from all the neural networks are  concatenated and used to train an MLP for interaction prediction. 


\section{Experiments and Results}
\subsection{Dataset details}
Three datasets, Binding DB, BRENDA, and KEGG, were used to evaluate CSI (Table \ref{tab:data_tab}A). Binding DB has the highest  number of compounds and the highest number of compounds per sequence (9.34 ratio), as expected from a drug-target interaction dataset. For the BRENDA dataset, we extracted interactions between enzymes and ligands as positive interactions. The listed inhibitor interactions were included as labeled negative interactions for the interaction predictor training \citep{visani2021enzyme}. For the KEGG dataset, the interactions were extracted from  reactions available in the KEGG database.  The two enzymatic datasets, the BRENDA and KEGG datasets, have overlap as the BRENDA database covers enzymes interacting with both natural and non-natural substrates, while the KEGG database covers natural interactions found in living organisms. The KEGG database provides detailed information on the underlying biochemical reactions, which enabled stratification on  interaction features. The two datasets have 757 compounds that had the same canonical SMILES. Of the 21,367 unique sequences in KEGG dataset, 10,948  are in the BRENDA dataset. The KEGG dataset has approximately $3 \times$ more interactions than the BRENDA dataset. 

For compound-based stratification (Table \ref{tab:data_tab}B), we report  the size of each strata.  Within each strata, the number of views is the square of the number of sequences divided by two as CMC is applied to pairs of sequences within each strata.  A large size therefore indicates  rich views within the strata. To assess the overlap among the strata, we report the average sharing among the strata and their Jaccard similarities. This latter metric gives a sense of how varied the views are across keys while also considering the strata size. 
We similarly summarize these metrics for sequence-based stratification (Table \ref{tab:data_tab}C). 
The KEGG dataset has more average shared sequences across compound keys (0.06) compared to the others, while the BindingDB dataset had more average shared compounds  across sequence keys (0.14).  When considering the Jaccard score, BindingDB has the highest similarities per strata for both compound- and sequence-based stratification.


For the interaction prediction task (Table \ref{tab:data_tab}D), the training data consists of positive examples comprising  protein-compound pairs that are known to interact. The negative examples are randomly selected compound-sequence pairs. The selection strategy of the negatives  reflects nature as most compounds and proteins do not interact. For training, we used a negative to positive ratio as 5:1, taking care to appropriately weight the loss during training. We created two kinds of test sets.  The \textbf{Test}  set included both positive and negative examples taken from the same distribution as the training set. We also generated test sets with 5X, 10X and 25X the number of negatives as positives to evaluate the impact of negative-to-positive ratio in  test. To test the generalizability of the model, we created an \textbf{Unseen Test} set that comprised the 5\% least frequent compounds and sequences in each dataset, which were held out from the training data set. We assume a 1:1 negative-to-positive ratio for the Unseen Test.

\subsection{Baseline Model}\label{baseline}
While our proposed data stratification strategy can be applied to any interaction model, we create a baseline model (Supplementary File, Section 1) that utilizes Graph Neural Networks (GNNs) to encode the molecules, and  Convolutional Neural Networks (CNNs) to encode the sequences.

\begin{table}[!t]
  \centering
  \fontsize{6.0pt}{6.0pt}\selectfont
  \caption{Statistics for the three evaluation datasets. (A) Base statistics. (B) Strata statistics when stratifying by compound. (C) Strata statistics when stratifying by sequence. (D)  The number of positive examples for  various data splits. }
    \begin{tabular}{clrrr}
\toprule
    \multicolumn{5}{c}{(A) Statistics for the interaction dataset } \\
    \midrule
    \multicolumn{2}{c}{} & \multicolumn{1}{c}{BindingDB} & \multicolumn{1}{c}{BRENDA} & \multicolumn{1}{c}{KEGG} \\
    \midrule
    \multicolumn{2}{l}{Number of interactions} & 68,347 & 40,693& 127,884  \\
    \multicolumn{2}{l}{Unique compounds} & 29,149 & 8,891 & 6,087  \\
    \multicolumn{2}{l}{Unique sequences} & 3,120  & 13,330& 21,367  \\
    \multicolumn{2}{l}{Compound-to-sequence ratio} & 9.34  & 0.67 & 0.28 \\
    
    \midrule
    \multicolumn{5}{c}{(B) Stratification on compounds (reported in sequences per strata)} \\
    \midrule
        \multirow{7}[0]{*}{\parbox{1.5cm}{}} & Average strata size & 2.18  & 7.2 & 21 \\
            & Standard deviation & 6.85 & 31.7 & 64.1 \\
    & Maximum strata size & 339 & 1518 & 2065 \\
    & Minimum strata size & 1 & 1 & 1 \\
    & Average sharing among strata & 0.01 & 0.01 & 0.06 \\
    & Average Jaccard similarity among strata & 0.004 & 0.001 & 0.001 \\
    \midrule
    \multicolumn{5}{c}{(C) Stratification on sequences (reported in compounds per strata)} \\
    \midrule
    
    \multirow{7}[0]{*}{\parbox{1.5cm}{}} & Average strata size& 20.4  & 3.5 & 5.9 \\
     & Standard deviation& 47.4 & 4.1 & 8.9 \\
    & Maximum strata size & 623 & 107 & 331 \\
    & Minimum strata size & 1 & 1 & 1 \\
    & Average sharing among strata & 0.14 & 0.04 & 0.06 \\
    & Average Jaccard similarity among strata & 0.003 & 0.001 & 0.001 \\

    \midrule
    \multicolumn{5}{c}{(D) Number of positives per data split} \\
    \midrule
   & {Training} &  49,746 & 32,536 & 100,999 \\
    
    &{Validation}  & 6,142 & 4,095 & 12,626 \\
   &{Test} & 7,833 & 4,925 & 14,261 \\
    &{Unseen Test}  & 1,609 & 885 & 1,636 \\
    \botrule
    \end{tabular}%
  \label{tab:data_tab}%
\end{table}%

\subsection{Experimental Setup}\label{expr_setup}
To evaluate the CSI model, we measure the model's performance in ranking positive examples ahead of negative examples, as well as the model's ability to rank a molecule or sequence that has the highest probability of interacting with a sequence or molecule respectively. We used Average Precision, Mean Average Precision and R-Precision as the metrics, the details of which are available in Section 2 of Supplementary material. 



The CSI model was trained in two steps.  In the contrastive learning step, the encoders generating $z_{v1}$ and $z_{v2}$ were trained using CMC on the congruent and non-congruent data views. The model was trained for 700 epochs. The best temperature $\tau$ was found to be 0.07 (we tried a range of 0.05-0.08). Adam \citep{kingma2014adam} was used as the optimizer. In the interaction prediction step, the training set was divided into training, validation and test sets in ratio 8:1:1. In this step, the  predictor model was trained for 200 epochs, with early stopping on validation loss. The optimizer used was Adam.

\subsection{Results on stratification by compounds and sequences}\label{res_comp_seq_strat}
The results for the three datasets are reported for test set with a negative-to-positive ratio of 1:1 (Table \ref{tab:results}A). The CSI model shows improved performance across all datasets and across all metrics. Specifically, AP is improved by 18.2\% on the BindingDB dataset, 39\% on the BRENDA dataset, and 
 13.7\% on the KEGG dataset over the baseline model. The improvement in MAP over baseline is maximum on the BindingDB dataset (23.8\%) for test data sorted by sequences while it is maximum on KEGG (26.2\%) for test data sorted by compounds. 
 For the Unseen Test set, CSI also shows improved performance across all metrics, and across all datasets (Table \ref{tab:results}B). Specifically, AP improvements are  2.6\%, 18.2\% and 1.6\% for BindingDB, BRENDA and KEGG datasets, respectively. 

The CSI model uses  embeddings learnt based on compound and sequence stratification. To determine which of the two stratification strategies contributes more to performance gains over the baseline model, we performed ablation studies on BindingDB (non-enzymatic) and KEGG (enzymatic) datasets (Table \ref{tab:results}C). Independently, compound and sequence stratification each contribute significantly to CSI's performance over the baseline. For BindingDB, with sequence based stratification alone, the AP drops from 0.992 (with both stratification) to 0.972. The drop is minimal (0.992 to 0.991) when using only compound-based stratification. These results indicate that for BindingDB, the compound based stratification contributes maximally to the CSI model's performance. For KEGG, the AP drops from 0.969 when using both stratification strategies to 0.953 when using only compound-based stratification. The drop is lesser when switching to sequence based stratification (0.969 to 0.960). This indicates that for KEGG, the sequence-based stratification contributes maximally to the CSI model's performance. The performance of the CSI model also scales well when the ratio of negative to positive examples in increased to mimic what happens in nature (Supplementary File, Section 3).


\begin{table*}[!t]
  \centering
  \fontsize{7.5pt}{7.25pt}\selectfont
  \caption{Interaction prediction results for a negative data ratio of 1:1 for the baseline and CSI models for the BindingDB, BRENDA and KEGG datasets. AP and R-Precision are reported for the entire dataset. MAP, mean R-Precision, MAP@3 and R-Precision@1 are reported for data sorted by compounds and by sequences.
  (A) Test set. (B) Unseen Test. (C) Ablation study to determine the individual contributions of each stratification strategy against using both strategies together. }
  


    \begin{tabular}{@{}p{5em}p{8.0em}cccccccccc@{}}
    \toprule

          & \multicolumn{1}{r}{} & \multicolumn{2}{c}{Overall} & \multicolumn{4}{c}{Compound}  & \multicolumn{4}{c}{Sequence} \\
          \cmidrule(lr){3-4}\cmidrule(lr){5-8}\cmidrule(lr){9-12}
          & \multicolumn{1}{r}{} & AP    & R-Precision & MAP   & R-Precision & Map@3 & Precision@1 & MAP   & R-Precision & Map@3 & Precision@1 \\
          \toprule     
               \multicolumn{12}{c} {(A) Test set}\\
          \midrule
    \multirow{2}{*}{BindingDB} & Baseline  & 0.839 & 0.783 & 0.914 & 0.844 & 0.913 & 0.842 & 0.802 & 0.709 & 0.799 & 0.681 \\
          & CSI  & 0.992 & 0.971 & 0.996 & 0.993 & 0.996 & 0.993 & 0.993 & 0.993 & 0.994 & 0.993 \\
          \midrule
    \multirow{2}[0]{*}{BRENDA} & Baseline  & 0.778 & 0.713 & 0.804 & 0.680  & 0.804 & 0.680  & 0.874 & 0.803 & 0.874 & 0.791 \\
          & CSI  & 0.991 & 0.970  & 0.995 & 0.990  & 0.996 & 0.993 & 0.978 & 0.993 & 0.978 & 0.975 \\
          \midrule
    \multirow{2}[0]{*}{KEGG} & Baseline  & 0.852 & 0.770  & 0.755 & 0.627 & 0.739 & 0.610  & 0.757 & 0.857 & 0.755 & 0.701 \\
          & CSI  & 0.969 & 0.902 & 0.953 & 0.918 & 0.956 & 0.939 & 0.809 & 0.968 & 0.808 & 0.793 \\
          \bottomrule

 \toprule
\multicolumn{12}{c} {(B) Unseen Test}\\
    \toprule
   \multirow{2}[0]{*}{BindingDB} & Baseline Model & 0.975 & 0.918 & 0.995 & 0.991 & 0.995 & 0.991 & 0.920 & 0.879 & 0.911 & 0.879 \\
          & CSI Model & 1.000 & 0.992 & 1.000 & 0.999 & 1.000 & 0.999 & 0.997 & 0.995 & 0.997 & 0.995 \\
          \midrule
    \multirow{2}[0]{*}{BRENDA} & Baseline Model & 0.845 & 0.703 & 0.934 & 0.891 & 0.927 & 0.891  & 0.901 & 0.857 & 0.897 & 0.841 \\
          & CSI Model & 0.999 & 0.985  & 1.000 & 1.000  & 1.000 & 1.000 & 0.982 & 1.000 & 0.982 & 0.982 \\
          \midrule
    \multirow{2}[0]{*}{KEGG} & Baseline Model & 0.873 & 0.771  & 0.779 & 0.682 & 0.753 & 0.676  & 0.730 & 0.963 & 0.730 & 0.718 \\
          & CSI Model & 0.886 & 0.773 & 0.915 & 0.878 & 0.908 & 0.869 & 0.718 & 0.934 & 0.717 & 0.697 \\
          \bottomrule

 \toprule
\multicolumn{12}{c} {(C) Ablation study}\\
    \toprule
    \multirow{2}[0]{*}{BindingDB} 
          & CSI Seq Strat & 0.972 & 0.921 & 0.982 & 0.973 & 0.978 & 0.969 & 0.971 & 0.960 & 0.983 & 0.982 \\
          & CSI Comp Strat & 0.991 & 0.971 & 0.987 & 0.989 & 0.996 & 0.991 & 0.983 & 0.982 & 0.982 & 0.982 \\
          \midrule
\multirow{2}[0]{*}{KEGG} 
          & CSI Seq Strat & 0.960 & 0.894 & 0.942 & 0.901 & 0.952 & 0.931 & 0.807 & 0.958 & 0.808 & 0.791 \\
          & CSI Comp Strat & 0.953 & 0.882 & 0.923 & 0.869 & 0.931 & 0.900 & 0.798 & 0.951 & 0.802 & 0.778 \\
          \bottomrule
 
    \end{tabular}%
  \label{tab:results}%
\end{table*}%

\subsection{Results on stratification by reaction features}
For the KEGG dataset, three interaction features were used to produce three  stratification strategies. The first strategy partitions the data based on enzymes catalyzing the same reaction (e.g., homologs).  The second strategy divides the interactions by the underlying  biotransformation pattern associated with the substrate-product pairs.  KEGG classifies reactions based on this property, and each class is referred to as an RCLASS \citep{kotera2014predictive}. Multiple reactions can belong to the same class and result in similar biotransformations.  The third strategy divides the interaction data by the Enzyme Commission (EC) number associated with the interaction.  EC numbers provide hierarchical classification on enzymes and are represented as four numbers separated by periods (e.g. L-lactate dehydrogenase is assigned EC number 1.1.1.27). Each such EC number is associated with one or more biochemical reactions. The three keys used to partition the KEGG interaction data are therefore: the reaction, RCLASS, and the EC numbers. Further details and analysis are provided in Supplementary File, Section 4.

The results are reported using AP (Table \ref{tab:KEGG_results}) as this metric was well correlated in earlier analysis with  other metrics. 
AP was reported for the baseline model on three datasets (1:1 negative-to-positive ratio, 5:1 ratio, and the Unseen Test) as well as for the CSI model for the same datasets. 
All stratification strategies yield improved results over the baseline model and stratification by compound and sequence across all test sets, where stratification by reaction outperforms the baseline by 
16.9\%, 62.6\% and 13\% on the 1:1, 5:1, and Unseen Test sets, respectively.  Comparing with stratification by compound and sequence, stratification by reaction yields  AP improvements over the compound and sequence stratification by 2.1\%, 6.3\% and 10.8\% on the 1:1, 5:1, and Unseen Test sets, respectively.



To evaluate how each of the views contributes to  improvements over the baseline, we perform an ablation study (Table \ref{tab:KEGG_results}B). As stratification by reaction resulted in higher performance over RCLASS- and EC-based stratification, the ablation study is applied to the reaction-based stratification model. 
The model was successively trained on each combination of two views (instead of all 3).
Removing $V_1$ (substrate-product view) contributed the most, when compared to removing the other two views, in reducing model performance, e.g., for the 5:1 positive-to-negative Test set, the AP performance is reduced from 0.963 to 0.751. The substrate-product view therefore contributes the most to the CSI model performance when stratifying by reaction features. We conjecture that the high similarity between substrate-product pairs contributes to higher mutual information when compared to the other views.

\begin{table}[!htbp]
\centering
\caption{AP results on stratification for the baselines (no stratification, compound/sequence stratification) and by the three interaction features: reaction, RCLASS, and EC. The three views, $V_1, V_2,$ and  $V_3$, correspond to substrate-product pairs, compounds-sequences and pairs of sequences. The ablation study considers only two of the views at a time.
} 
{\begin{tabular}{@{}cccc@{}}\toprule Model &
Test & Test(5:1) & Unseen Test\\\midrule
\hline 
    \multicolumn{4}{c}{(A) Summary of prior results} \\
\hline 
Baseline (no stratification) & 0.852 & 0.587 & 0.873\\
Compound/sequence stratification & 0.969 & 0.906 & 0.886\\
\hline 
    \multicolumn{4}{c}{(B) Interaction features} \\
\hline 
Reaction  ($V_1$, $V_2$, $V_3$) & 0.989 & 0.963 & 0.982\\
RCLASS  ($V_1$, $V_2$, $V_3$) & 0.990 & 0.913 & 0.954\\
EC  ($V_1$, $V_2$, $V_3$) & 0.988 & 0.943 & 0.979\\
\hline 
  \multicolumn{4}{c}{(C) Ablation study}\\
\hline
Reaction Strat ($V_1$, $V_2$) & 0.980 & 0.874 & 0.941\\
Reaction Strat ($V_2$, $V_3$) & 0.962 & 0.751 & 0.904\\
Reaction Strat ($V_1$, $V_3$) & 0.983 & 0.902 & 0.947\\\botrule
\end{tabular}\label{tab:KEGG_results}}{}
\end{table}

\vspace{-.25in}
\section{Conclusion}

CSI is a generalizable data stratification technique that exploits relationships among interacting objects to stratify training data and to define congruent (and non-congruent) views on one object in terms of the other.  Alternatively, CSI can exploit features of the interaction, and partitions the training data to generate such views. Paired with CMC,  CSI  learns representations that maximize the mutual information among congruent views, thus leading to enhanced representations for downstream interaction prediction task. We applied CSI to three compound-protein sequence datasets. We demonstrated significant improvements in interaction prediction performance across multiple metrics and across drug-target and enzymatic datasets when compared to a baseline model that does not exploit stratification.
We further demonstrated that, for our datasets, stratification by interaction features results in improved performance over stratification on object relationships.

\section*{Acknowledgements}


\section*{Funding}
 This research is supported by NSF Award 1909536. This research is also  supported by the NIGMS of the National Institutes of Health, Award R01GM132391.  The content is solely the responsibility of the authors and does not represent the  views of the NSF nor the NIH.

\bibliographystyle{package/natbib}
\bibliography{references}

\end{document}


\title{\textbf{CSI}: \underline{\textbf{C}}ontrastive Data \underline{\textbf{S}}tratification for \underline{\textbf{I}}nteraction Prediction and its Application to Compound-Protein Interaction Prediction
\textbf{Supplementary File}}
\date{}

\author[1]{Apurva Kalia}
\author[2]{Dilip Krishnan}
\author[1,3]{Soha Hassoun}
\affil[1]{Department of Computer Science, Tufts University, Medford, MA 02155, USA}  
\affil[2]{Google Research}
\affil[3]{Department of Chemical and Biological Engineering, Tufts University, Medford, MA 02155, USA.}

\maketitle
\section{Details of Baseline Model}
\begin{figure*}[!h]
    \centering
    \includegraphics[width=4.5in]{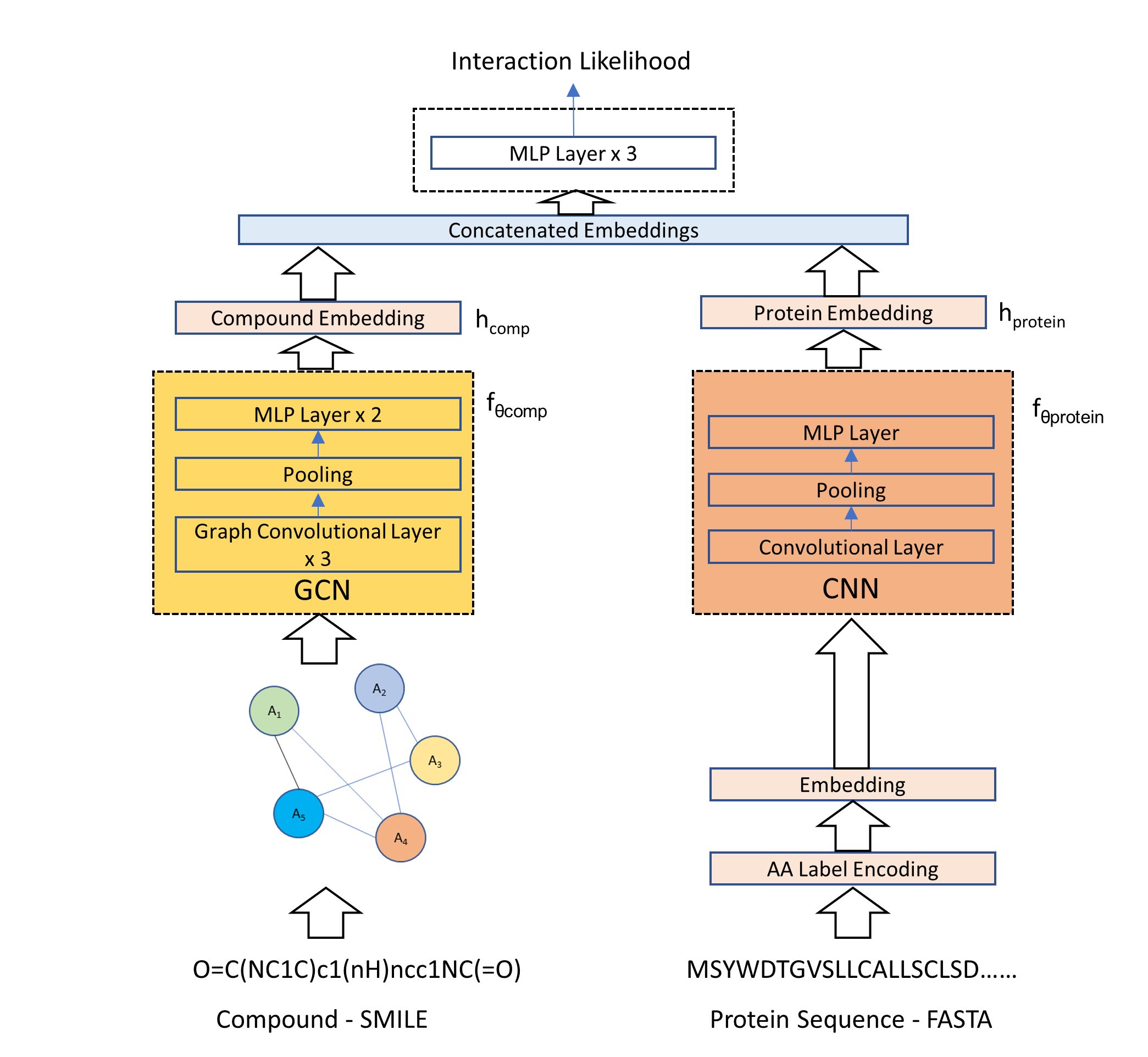}
    \caption{Compound-protein interaction prediction model used as the baseline. Interaction likelihood is predicted based on learned molecular and protein interactions. 
    }
    \label{fig:baseline}
\end{figure*}

Our baseline model is based on GraphDTA  \citep{nguyen2021graphdta}. GraphDTA offers a simple and generalizable method to create graph-based encoders for molecules represented in SMILES and CNN- based encoders for sequences and achieves 10-15\% improved results compared to other models \citep{ozturk2018deepdta, ozturk2019widedta, he2016simboost, cichonska2017computational}, and thus is a strong baseline. The baseline model architecture comprises of encoders for molecules and sequences, followed by MLP layers for interaction prediction (Figure \ref{fig:baseline}). Compounds represented in SMILES format are converted to to a molecular graph using rdkit \citep{landrum2013rdkit}. For our baseline, we use node features as the atom type, atomic mass, valence, is atom in ring, formal charge, radical electrons, chirality, degree, number of hydrogens and aromaticity. Bond features  are the bond type, whether the bond is part of a ring, conjugacity and one hot encoding of the stereo configuration of the bond. Compound embeddings are learned using a multi-layer Graph Neural Network (GNN) encoder. The network consists of Graph Convolutional Networks (GCNs) \citep{kipf2016semi} that aggregate information at each node. The GCNs are followed by a pooling layer and two fully connected layers. Each amino acid within protein sequences (in FASTA format) is first converted to a numeric code used to generate learnable embeddings. Sequence embeddings are passed to a protein encoder which consists of a 1-d Convolutional Neural Network (CNN) followed by a pooling layer and a fully connected layer. The compound and sequence embeddings are  concatenated  for the final interaction likelihood prediction. The final predictor is a 3-layer MLP with the first two layers each reducing the embedding dimensionality by half and the final layer making a binary prediction. Importantly, the architecture of the GCN and CNN encoders of the baseline model are used for CSI to ensure a fair comparison between the CSI and baseline model.

\section{Metrics used for model performance}
To measure the model performance, it is important to choose metrics that can measure the model's ability to discriminate between positive and negative examples without needing to define a threshold dividing the positive from negative cases. This is important because the threshold could vary for different molecule-enzyme combinations. We used the following metrics to measure model performance:
\begin{itemize}
    \item \textbf{Average Precision (AP)} is measured across each dataset, reflecting the model's ability to distinguish positive and negative examples. 
    \item \textbf{R-precision}, also measured across each dataset, measures the model's ability to accurately predict the R known positive interactions.
     \item \textbf{Mean Average Precision (MAP)} measures the the AP per compound (or per protein sequence) averaged over interactions sorted by compound (or protein sequence), thus indicating the model's ability to predict the likelihood of interaction for a given  compound (or protein sequence). 
    \item \textbf{MAP@3} reports MAP on the top 3 ranked items i.e the top 3 sequences per compound, or the top 3 compounds per sequence.
    \item \textbf{Precision@1} measures the ability of the model to correctly predict a top ranked interacting item.
\end{itemize}

\begin{table}[!h]
  \centering
  \caption{Statistics for the KEGG dataset for three different stratification strategies by interaction features. We report the total number of objects in each view with each stratification strategy, the average number of objects in each views over all keys, as well as the distribution of objects in each view.}
    \begin{tabular}{lcccc}
    \multicolumn{1}{c}{} &  
    \multicolumn{1}{c}{Number of views} & 
    \multicolumn{1}{c}{Mean objects} &
    \multicolumn{1}{c}{std-dev objects} &
    \multicolumn{1}{c}{Max objects} \\
    \midrule
    \multicolumn{5}{c}{(A) Stratification on reaction. Number of keys is 6,059} \\
    \midrule
    \multicolumn{1}{l}{$V_1$(comp-comp)} & 9,091 & 1.50 & 0.58 & 4 \\
    \multicolumn{1}{l}{$V_2$(comp-seq)} & 124,046 & 20.47 & 35.32 & 387 \\
    \multicolumn{1}{l}{$V_3$(seq-seq)} & 52,759 & 8.71 & 15.36 & 129 \\
    \midrule
    \multicolumn{5}{c}{(B) Stratification on RCLASS. Number of keys is 2,158} \\
    \midrule
    \multicolumn{1}{l}{$V_1$(comp-comp)} & 5,765 & 2.67 & 7.09 & 134 \\
    \multicolumn{1}{l}{$V_2$(comp-seq)} & 172,382 & 79.88 & 441.91 & 15635 \\
    \multicolumn{1}{l}{$V_3$(seq-seq)} & 50,723 & 23.50 & 107.47 & 2994 \\
    \midrule
    \multicolumn{5}{c}{(C) Stratification on EC. Number of keys is 3,363} \\
    \midrule
    \multicolumn{1}{l}{$V_1$(comp-comp)} & 8,493 & 2.53 & 3.00 & 79 \\
    \multicolumn{1}{l}{$V_2$(comp-seq)} & 98,254 & 29.22 & 170.72 & 7500 \\
    \multicolumn{1}{l}{$V_3$(seq-seq)} & 17,563 & 5.22 & 8.47 & 129 \\
    \end{tabular}%
  \label{tab:data_kegg}%
\end{table}%

\section{Scaling of model performance with increasing negative to positive ratio}
\begin{figure*}[t]
    \centering
    \includegraphics[width=\linewidth]{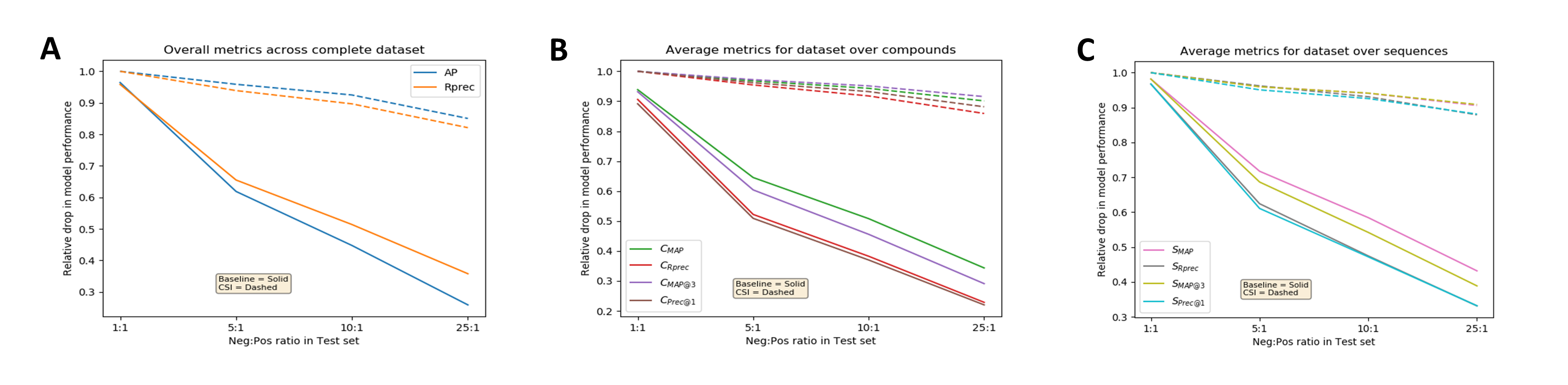}
    \caption{Model performance evaluation for various negative-to-positive ratios in the Test set. (A) AP and R-precision trends for various negative-to-positive ratios for Test set. (B) MAP, mean R-Precision, MAP@3, R-Precision@1 trends for Test set interactions sorted by compounds. (C) MAP, mean R-Precision, MAP@3, R-Precision@1 trends for Test set interactions sorted by sequences.}
    \label{CSIfig:result_scale}
\end{figure*}

We measured model performance scaling with respect to the negative-to-positive ratio (Figure \ref{CSIfig:result_scale}). Assuming the performance with 1:1 ratio to be 1.0, baseline model AP performance drops by 73\% when using 25:1 ratio. Meanwhile, the performance of the CSI model drops only by 18\% at the 25:1 ratio. For metrics measured per compound, the MAP metric drops by 75\% for the baseline model for the 25:1 ratio whereas the same metric drops by only 14\% for the CSI model. For metrics measured by sequence, the drops in MAP are 66\% and 13\% for the baseline and CSI model respectively. These results indicate that CSI performs better than the baseline model at predicting the negatives correctly. Further, the positives and negatives continue to be well separated even as the ratio of negatives increases.

\section{Details of KEGG stratification by reaction features}
An enzymatic dataset like KEGG can also be stratified by reaction features - reaction, RCLASS and EC numbers. For each strata (Table \ref{tab:data_kegg}), three different views of the data are possible: substrate-product pairs, compounds-sequence pairs and pairs of sequences.  The number of keys per strategy differ, where stratification on reactions provides the most number of keys. The key choices subsequently affect the total number of views and the size of each strata within the views. Regardless of the key, there are more compound-sequence views ($V_2$) than the other two views, and fewest compound-compound views ($V_1$). We examine the strata to determine if any one particular reaction consistently contributed to the maximum strata size.  The largest compound-sequence partition under the reaction stratification strategy is due to the ammonia-ubiquinol reaction (KEGG reaction R00148), which contributes to  nitrogen metabolism.
This reaction is catalyzed by ammonia mono-oxygenase (EC 1.14.99.39), which is present in 46  organisms leading to many sequences for the same enzyme. For RCLASS-based stratification, the largest compound-sequence partition is for RC00001 which is part of glutathione metabolism. This basic reaction class encompasses 14 different reactions, catalyzed by 18 different EC classes - leading to a large number of compound-sequence pairs. For EC-based classification, the largest compound-sequence partition is for glutathione transferase (EC 2.5.1.18), which catalyzes 24  reactions, and present in 423 different organisms.

For training, validation and testing on the interaction features, the methodology that was followed is similar to our prior strategy (Section 3.3 in the paper). Training, validation and test were split in 8:1:1 ratio. In addition, a test set with 5x more negatives than positives as well as the Unseen Test set were created.

\bibliographystyle{package/natbib}
\bibliography{references}